%% file: main.tex
\documentclass[11pt]{article}

\input{UTILS/PreSetting}

\usepackage{tabularx}
\usepackage{ragged2e}

\begin{document}

\newcommand{\cgexec}{\texttt{executability}\xspace}
\newcommand{\cgcorrect}{\texttt{correctness}\xspace}
\newcommand{\datasetname}{\texttt{BioDSBench}\xspace}


\title{\LARGE \textbf{Can Large Language Models Replace Data Scientists in Biomedical Research?}}


\author[1,+]{Zifeng Wang}
\author[1,+]{Benjamin Danek}
\author[2]{Ziwei Yang}
\author[3]{Zheng Chen}
\author[1,4,\#]{Jimeng Sun}

\affil[1]{School of Computing and Data Science, University of Illinois Urbana-Champaign, Urbana, IL, USA}
\affil[2]{Bioinformatics Center, Institute for Chemical Research, Kyoto University, Kyoto, Japan}
\affil[3]{Institute of Scientific and Industrial Research, Osaka University, Osaka, Japan}
\affil[4]{Carle Illinois College of Medicine, University of Illinois Urbana-Champaign, Champaign, IL, USA}

\affil[+]{\em{Equal contribution}}
\affil[$\#$]{\em{Correspondence: \href{mailto:jimeng@illinois.edu}{jimeng@illinois.edu}}}

\date{}

\maketitle



\abstract{Data science plays a critical role in biomedical research, but it requires professionals with expertise in coding and medical data analysis. Large language models (LLMs) have shown great potential in supporting medical tasks and performing well in general coding tests. However, existing evaluations fail to assess their capability in biomedical data science, particularly in handling diverse data types such as genomics and clinical datasets. To address this gap, we developed a benchmark of data science coding tasks derived from the analyses of 39 published studies. This benchmark comprises 293 coding tasks (128 in Python and 165 in R) performed on real-world TCGA-type genomics and clinical data. Our findings reveal that the vanilla prompting of LLMs yields suboptimal performances due to drawbacks in following input instructions, understanding target data, and adhering to standard analysis practices. Next, we benchmarked six cutting-edge LLMs and advanced adaptation methods, finding two methods to be particularly effective: chain-of-thought prompting, which provides a step-by-step plan for data analysis, which led to a 21\% code accuracy improvement (56.6\% versus 35.3\%); and self-reflection, enabling LLMs to refine the buggy code iteratively, yielding an 11\% code accuracy improvement (45.5\% versus 34.3\%). Building on these insights, we developed a platform that integrates LLMs into the data science workflow for medical professionals. In a user study with five medical professionals, we found that while LLMs cannot fully automate programming tasks, they significantly streamline the programming process. We found that 80\% of their submitted code solutions were incorporated from LLM-generated code, with up to 96\% reuse in some cases. Our analysis highlights the potential of LLMs to enhance data science efficiency in biomedical research when integrated into expert workflows.}

\newpage

\section*{Introduction}\label{sec:intro}
In biomedical research, data science plays a pivotal role in analyzing complex datasets, such as clinical trial data and real-world data (RWD), which are critical to improving patient care and advancing evidence-based medicine~\cite{radenkovic2019data}. For example, it was reported that for a pharmaceutical company, the insights derived from the RWD analysis could unlock up to \$300M values annually by optimizing the design and execution of clinical trials~\cite{davidmckinsey2024}. Data scientists, at the core of this process, require years of coding experience, along with a deep understanding of various types of medical data, including clinical data from patients and omics data, while also collaborating closely with medical professionals~\cite{lisaharvard2024}. However, the growing demand for data science skills and the limited availability of experienced data scientists have become bottlenecks in the biomedical research process~\cite{meyer2019healthcare}. Coding is central to the work of data scientists, underpinning essential tasks such as statistical modeling, data cleaning, and visualization using Python and R. Given this, exploring methods for streamlining the coding process in biomedical data science is crucial to accelerate drug development and improve patient outcomes.

Code generation has been extensively studied with large language models (LLMs), which have demonstrated strong capabilities in tasks like code completion~\cite{chen2021evaluating}. Continuous efforts have been made to develop more powerful code-specific LLMs~\cite{li2022competition,luowizardcoder,lozhkov2024starcoder}, refine prompting strategies~\cite{zhang2023repocoder}, integrate external knowledge through retrieval-augmented generation (RAG)~\cite{parvez2021retrieval,wang2024coderag}, and enable LLMs to self-reflection~\cite{chenteaching2024}. These advancements lead to the LLM-based platform for software development~\cite{wang2024opendevin} and data analysis~\cite{zheng2024opencodeinterpreter}. Although LLMs have been evaluated for general programming tests~\cite{chen2021evaluating,austin2021program,hendrycks2measuring,liu2024your}, software engineering~\cite{jimenezswe2024}, and data analysis ~\cite{huang2022execution,lai2023ds}, assessments specifically targeting biomedical data science remain scarce. In medicine, recent work has introduced LLMs to automate machine learning modeling~\cite{tayebi2024large} and support bioinformatic tool development~\cite{tang2024biocoder}, but they do not cover broad data science tasks or assess to what extent LLMs can collaborate with humans to complete programming tasks. Therefore, this paper seeks to build a comprehensive code generation dataset to assess to what extent cutting-edge LLMs can automate biomedical data analysis, modeling, and visualization.

Our objective is to evaluate the practical utility of LLMs in handling complex biomedical data and performing associated data science tasks. To this end, we identified 39 medical and biomedical studies published in medical journals that were linked to patient-level datasets (Fig.~\ref{fig:results_1}a). A list of example studies and data from the linked patients can be found in Extended~Fig.~\ref{fig:example_study}. We started by extracting and summarizing the analyses performed in these studies, such as patient characteristic exploration and Kaplan-Meier curves. We then developed the code necessary to reproduce these analyses and the results reported in these studies. These coding tasks, along with their reference solutions, were crafted and cross-verified manually to ensure accuracy. The result was a collection of 293 diverse and high-quality data science tasks, covering primary tools used in Python and R, e.g., \texttt{lifelines} for survival analysis in Python and \texttt{Bioconductor} for biomedical data analysis in R. Additionally, we categorized the difficulty of these tasks into Easy, Medium, and Hard, by the number of ``semantic lines" of code in the reference solutions (Fig~\ref{fig:results_1}d). The semantic lines metric aggregates lines of code that serve the same operation into a single unit, providing a clear measure of task complexity. A detailed overview of the dataset and its characteristics is provided in Fig.~\ref{fig:results_1}.

In this work, we evaluated the extent to which LLM can automate biomedical data science tasks. We benchmarked six state-of-the-art LLMs using various methods, including chain-of-thought prompting, few-shot prompting, automatic prompting, self-reflection, and retrieval-augmented generation (Fig.~\ref{fig:results_1}f). Our analysis focused on both the accuracy and quality of the code generated by these models. Although we found that current LLMs are not yet capable of fully automating complex biomedical data science tasks, they do generate code that is highly similar to the correct final solutions. Building on this insight, we investigated the development of an integrated platform designed to facilitate collaboration between human experts and artificial intelligence (AI) to streamline data science coding tasks. This platform aims to enhance the productivity of biomedical researchers by integrating LLMs into established data science workflows, with a focus on user-friendliness and the reliability of the outputs. Our results demonstrated that the platform significantly improved human experts' efficiency in executing data science tasks, highlighting the promising future of human-AI collaboration in biomedical data science.

\section*{Results}\label{sec:results}

\subsection*{Creating data science tasks from medical and biomedical studies}
\begin{figure}[htbp]
    \centering
    \includegraphics[width=0.9\linewidth]{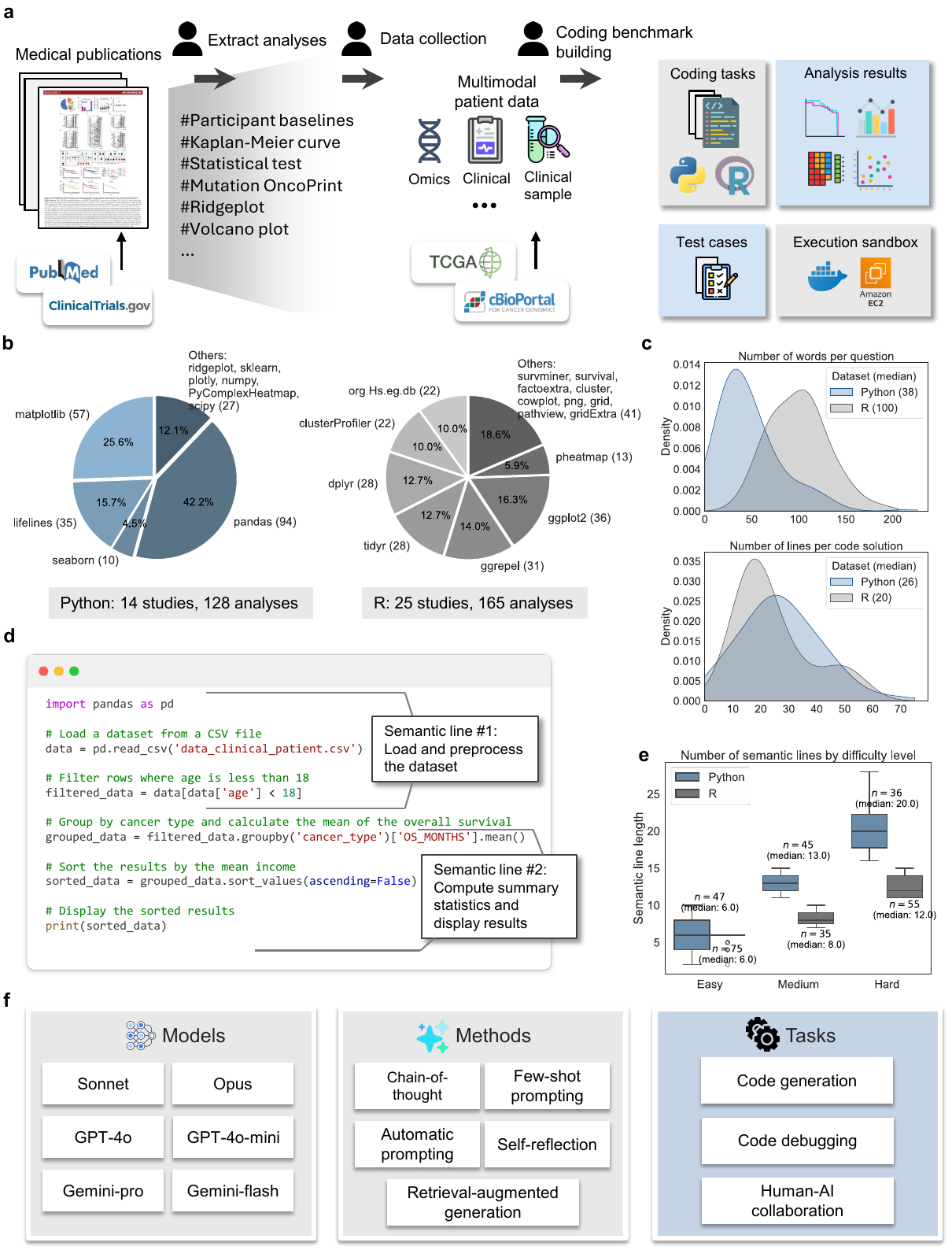}
    \caption{\textbf{Framework overview.} \textbf{a}, we created a data science coding dataset based on the extracted analyses from medical publications. \textbf{b}, the total number of analysis tasks and studies in the testing data, which also covers a diverse set of tools and libraries. \textbf{c}, illustration of the complexity of the tasks by the distributions of question length and answer length. \textbf{d}, an example of semantic lines. \textbf{e}, the distribution of semantic lines in the reference answers across different difficulty levels. \textbf{f}, the selected models, adaptation methods, and coding tasks in this study.}
    \label{fig:results_1}
\end{figure}

We curated our testing dataset, \datasetname, to reflect the real-world challenges that data scientists face in biomedical research. The dataset is grounded in published medical studies and linked to patient-level datasets from cBioPortal~\cite{cbioportal}. These datasets are diverse, each containing data from hundreds to thousands of patients, including clinical information such as demographics and survival data, lab results, and omics data like gene expression, mutations, structural variants, and copy number alterations. Unlike prior studies that mostly focus on single tables, each study in our dataset is linked to multiple tables and each can be with thousands of columns, providing a more complex and realistic basis for evaluating data science workflows. This setup mirrors the multifaceted nature of real-world biomedical research, where data scientists must integrate and analyze information from various sources to generate insights. The dataset building and evaluation framework is illustrated in Fig.~\ref{fig:results_1}a.

We obtained the publications associated with these datasets from PubMed and extracted the types of analysis in these papers. Through this process, we identified common analyses frequently performed in biomedical research, such as summarizing patient baseline characteristics, plotting Kaplan-Meier curves to assess treatment effects across groups, and creating mutational OncoPrints to highlight significant gene mutations in specific patients. This extraction process allowed us to filter and refine the initial set of studies, ensuring both diversity and comprehensiveness in the analyses covered. After this refinement, we retained 39 studies that were used to create the final testing dataset.

We designed a series of coding questions in a step-by-step format, mirroring the logical progression of the analyses in the original studies, ultimately leading to the main findings. For example, a study may include exploratory data analysis, gene mutation analysis to detect abnormal mutation patterns, survival analysis to visualize patient outcomes, and statistical tests to verify significance. Correspondingly, we developed coding questions for each step, ensuring that earlier steps provide the necessary groundwork for subsequent analyses. As such, each coding task consists of five components: (1) the input question, (2) dataset schema description, (3) prerequisite code (called ``prefix''), (4) reference solutions, and (5) test cases. This design reflects the practical setup that data scientists encounter in real-world projects. In total, we manually curated 128 analysis tasks in Python and 165 in R based on the extracted analyses from 39 studies. As shown in Fig.~\ref{fig:results_1}c, the input questions typically consist of 50-100 words describing task and output requirements, while reference code solutions span more than 20 lines, sometimes exceeding 50 lines, reflecting the complexity of the tasks.

We evaluated the difficulty of each coding task by calculating the number of semantic lines in the reference solutions. A semantic line aggregates multiple lines of code that contribute to the same operation, as illustrated in Fig.~\ref{fig:results_1}d. This approach prevents the difficulty assessment from being skewed by repetitive or tedious tasks that are fundamentally simple. The statistics for semantic lines and difficulty levels are presented in Fig.~\ref{fig:results_1}e. Our analysis shows that Python solutions tend to be more complex than R solutions, particularly for Medium and Hard tasks. This is largely due to R's rich ecosystem of medical-specific packages, which often allow for more direct solutions. In contrast, Python frequently requires additional customization and manual coding to achieve similar outcomes, contributing to higher complexity in Python coding tasks. The pie charts in Fig.~\ref{fig:results_1}b show the libraries frequently used in reference solutions.

\subsection*{LLMs are not yet ready for fully automated data science}
As illustrated in Fig.~\ref{fig:results_1}f, our evaluation framework consists of three key components: models, methods, and tasks. For the first component, we selected six cutting-edge LLMs: GPT-4o~\cite{gpt4o}, GPT-4o-mini~\cite{gpt4omini}, Sonnet~\cite{sonnet35}, Opus~\cite{opus3}, Gemini-pro~\cite{reid2024gemini}, and Gemini-flash~\cite{reid2024gemini}. These models represent a diverse set of advanced generalist LLMs capable of performing code generation based on input instructions. To explore their effectiveness and potential room for improvement in biomedical data science tasks, we applied several adaptation methods: chain-of-thought~\cite{wei2022chain}, few-shot prompting~\cite{10.5555/3495724.3495883}, automatic prompting~\cite{khattab2023dspy}, self-reflection~\cite{chenteaching2024}, and retrieval-augmented generation (RAG)~\cite{lewis2020retrieval}. 

For evaluation, we assessed the combinations of models and adaptation methods across three tasks: code generation, code debugging, and human-AI collaboration. The first two tasks were used to measure the models' accuracy in automating the coding process. We used Pass@$k$ as the primary metric, where $k$ represents the number of attempts the model is allowed to make to solve a coding task. This metric behaves like the probability that at least one out of the $k$ attempts is correct. Specifically, we selected $k=1$ as a strict benchmark to evaluate how well LLMs can automate tasks on the first attempt, providing insight into their immediate accuracy. Moreover, employing $k=5$ served as a broader metric, enabling us to examine the model's capability to improve with several attempts, thereby providing a more comprehensive evaluation of its potential to produce accurate solutions when given additional opportunities.

\begin{figure}[htbp]
    \centering
    \includegraphics[width=0.9\linewidth]{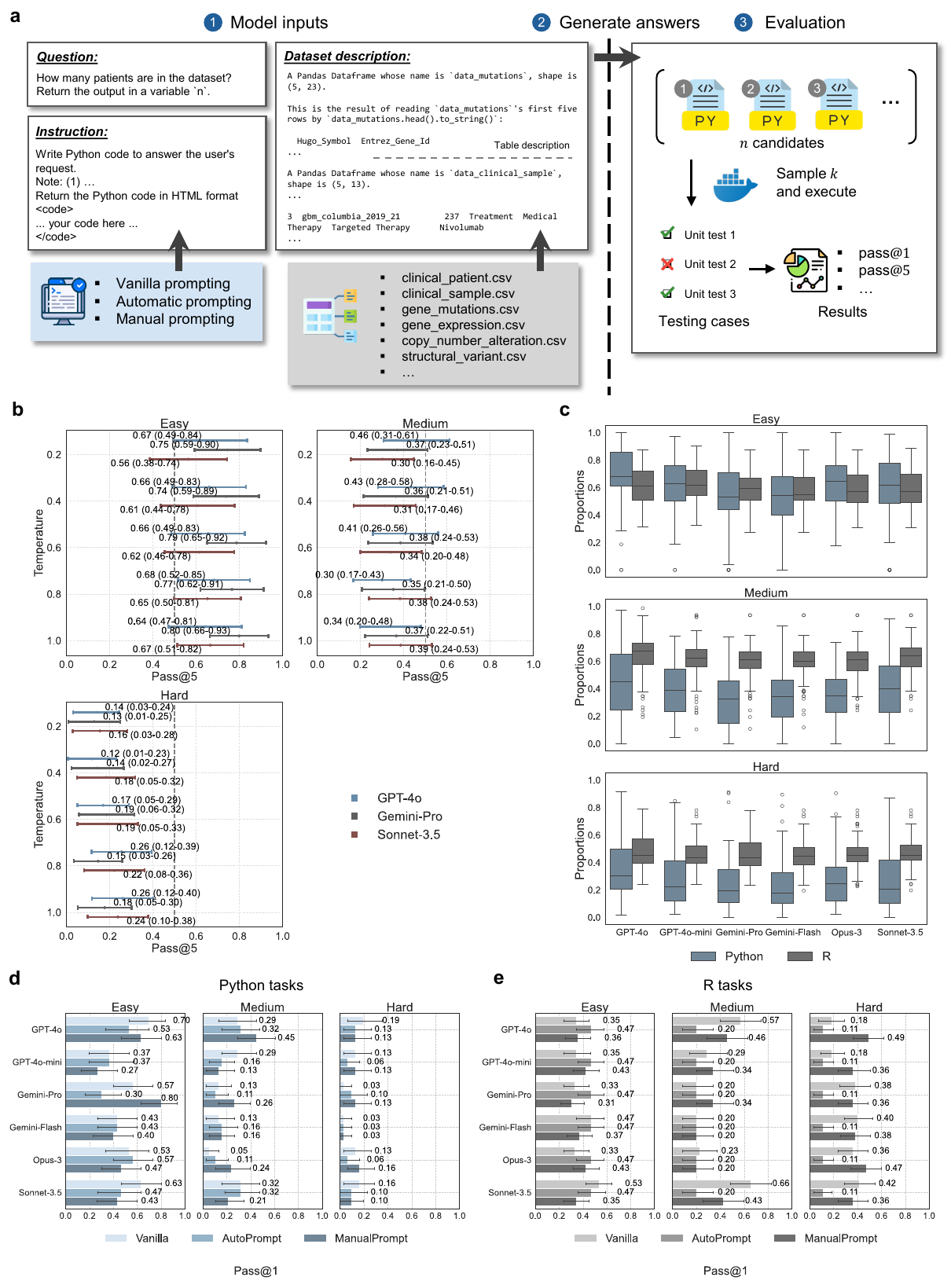}
    \caption{\textbf{Assessment of different models and adaptation methods in automating biomedical data science tasks.} \textbf{a}, the inputs for LLMs to generate the code and the associated evaluation process. \textbf{b}, the pass@5 of three LLMs with varying temperatures across difficulty levels in the Python coding dataset. \textbf{c}, the proportions of the reference solution code that can be drawn directly from the LLM-generated code. \textbf{d} and \textbf{e} show the pass@1 of six LLMs across difficulty levels in Python and R coding datasets, respectively.}
    \label{fig:results_2}
\end{figure}

We first evaluated the immediate accuracy of LLMs in generating code solutions on their initial attempt. For each task, the LLM is provided with a raw question that describes the target task, as well as a dataset description (Fig.~\ref{fig:results_2}a). The dataset description includes details such as table names, column names, and common cell values, which guide the LLM in identifying the correct table, column, and values to work with. Additionally, the instruction section offers supplementary guidance for the LLM during code generation. We used three types of instructions: Vanilla, Manual, and Automatic. The Vanilla instruction provides minimal guidance, merely instructing the LLM to solve the task, while Manual and Automatic instructions are more detailed, either manually crafted or optimized through automatic prompt generation~\cite{khattab2023dspy}. The generated code solutions must pass all the testing cases to be considered right.

From our experiments, we found that the vanilla adoption of LLMs cannot consistently produce perfect code for biomedical data science tasks across all difficulty levels. As shown in Fig.~\ref{fig:results_2}d, for Python tasks, the Pass@1 scores vary significantly based on task difficulty. For Easy tasks, most LLMs achieve Pass@1 rates in the range of 0.40-0.80. However, for Medium tasks, the Pass@1 rates drop to 0.15-0.40, and for Hard tasks, they range from 0.05 to 0.15. Performance differences also exist between different LLMs, particularly within the same series. For instance, the lightweight variant GPT-4o-mini generally underperforms compared to its larger counterpart, GPT-4o, with differences in performance of up to twofold in many cases. This highlights the limitations of current LLMs, especially as task complexity increases. The trend is similar in R, where performance declines with increased task difficulty, though there is a significant difference in performance between Python and R tasks (Fig~\ref{fig:results_2}e).

Diving deeper into the variations across instruction types, we observed that (1) in Python tasks (Fig.~\ref{fig:results_2}d), neither automatically generated prompts nor manually crafted prompts consistently outperformed the Vanilla prompts. For example, Vanilla performed better than AutoPrompt in 4 out of 6 LLMs for Easy tasks, 2 out of 6 for Medium tasks, and 4 out of 6 for Hard tasks. A similar trend was observed for R tasks (Fig.~\ref{fig:results_2}e). (2) More powerful models, such as GPT-4o and Gemini-Pro, showed greater benefits from carefully crafted instructions, particularly in Easy and Medium Python tasks. In contrast, lighter models like GPT-4o-mini and Gemini-Flash did not exhibit such improvements, and in some cases, complex instructions even seemed to hinder performance. This suggests that lightweight models may struggle to fully interpret and utilize complex instructions, which can reduce their effectiveness in data science coding tasks.

We adjusted the temperature settings to sample multiple solutions from LLMs and calculated Pass@5 scores for Python tasks (Fig.~\ref{fig:results_2}b). In most cases, increasing the temperature allows LLMs to generate more creative and diverse solutions, resulting in higher probabilities of producing a correct solution. This trend was consistent across all models, suggesting the potential benefit of having LLMs brainstorm multiple solutions to reach better outcomes. On average, LLMs solved more tasks when given five attempts compared to just one, as measured by Pass@1. However, despite this improvement, the overall performance remains far from perfect.

\subsection*{Unlocking the power of LLMs through strategic adaptations}
\begin{figure}[htbp]
    \centering
    \includegraphics[width=0.9\linewidth]{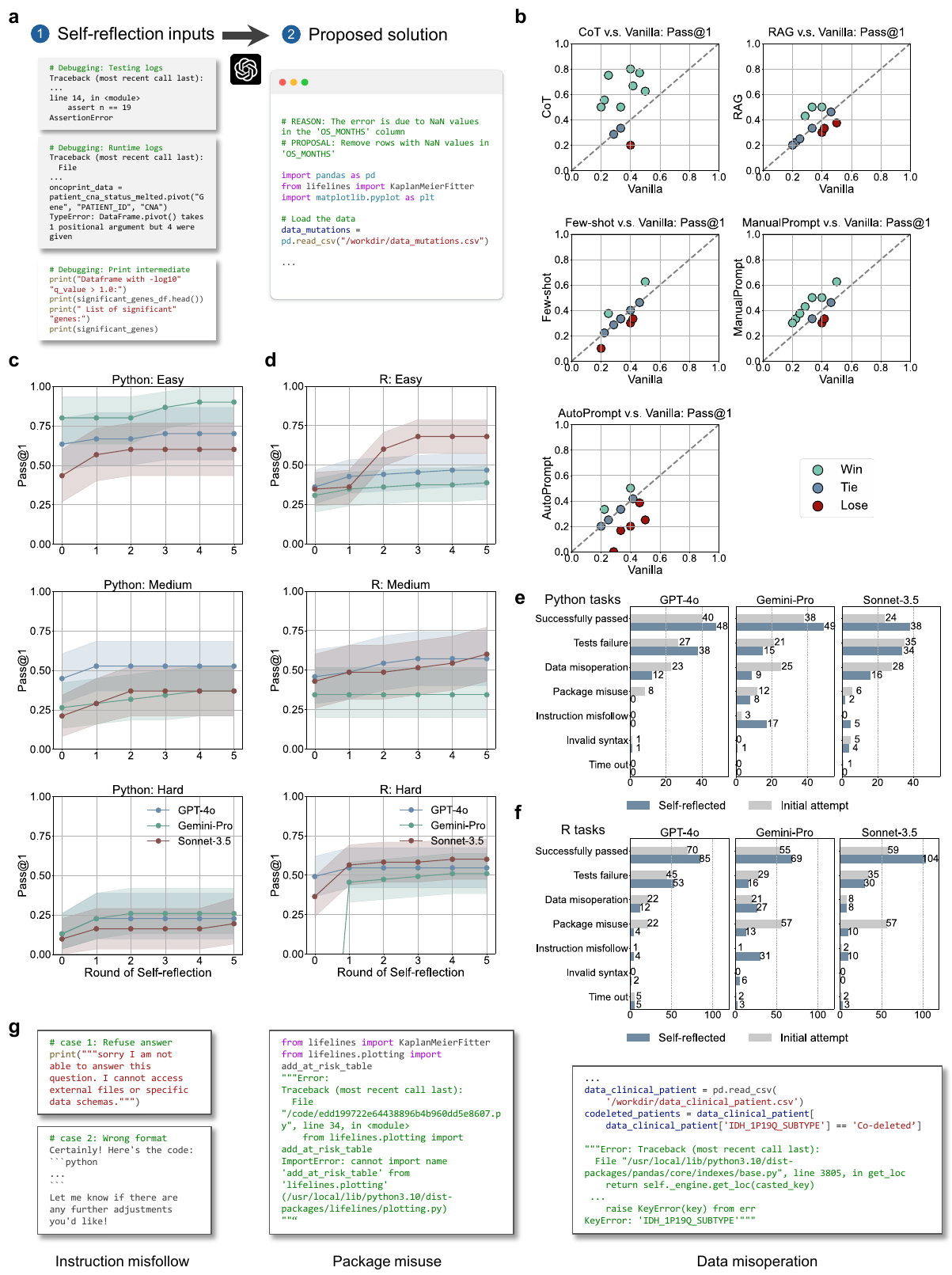}
    \caption{\textbf{Exploration of strategic adaptations and their effectiveness}. \textbf{a}, the inputs for LLMs' self-reflection are the testing logs, runtime logs, and the printing statements, from the initial code, and outputs the proposed solutions. \textbf{b}, study-level comparison of different adaptations versus vanilla methods. \textbf{c} and \textbf{d}, the Pass@1 with increasing rounds of self-reflections for Python and R tasks, respectively. \textbf{e} and \textbf{f}, the outcome classifications of code solutions before and after self-reflection for Python and R tasks, respectively. \textbf{g}, demonstrations of three error types.}
    \label{fig:results_3}
\end{figure}

Motivated by the varied performances of LLMs with different instruction levels, we hypothesized that tailored adaptations for LLMs in biomedical data science could lead to greater improvements. To test this, we introduced two key dimensions of adaptation: (1) enhancing LLM inference and reasoning by incorporating advanced instructions or external knowledge, and (2) employing multiple rounds of trial-and-error, allowing LLMs to iteratively correct their errors. The results of these adaptations are shown in Fig.~\ref{fig:results_3}.

For the first dimension of adaptation, in addition to Manual (ManualPrompt) and Automatic prompt optimization (AutoPrompt), we introduced three additional strategies: chain-of-thought (CoT), few-shot prompting (Few-shot), and retrieval-augmented generation (RAG). ManualPrompt incorporates human knowledge into the instructions, offering additional hints such as common error cases, key columns like unique patient identifiers, and specific guidance for certain analyses. AutoPrompt utilizes the DSPy prompt optimizer~\cite{khattab2023dspy}, which generates prompts via an LLM and selects the best one. We optimized prompts for Python code generation using three studies from the training set, keeping 11 studies as the test set. RAG equips LLMs with a Google search engine, enabling them to look up package documentation, StackOverflow discussions, and clinical knowledge before generating code solutions. Few-shot prompting adds several example question-and-answer pairs from the training set to guide the model. For CoT, we enriched the instructions with step-by-step guidance, asking the LLM to follow concrete steps toward the final solution. These instructions were manually created to ensure accuracy, mimicking scenarios where proficient data scientists provide more detailed input.

The comparison of adaptation strategies based on GPT-4o is illustrated in Fig.~\ref{fig:results_3}b. Each data point represents the average Pass@1 score achieved for coding tasks in a given study. A point on the diagonal line indicates equivalent performance between the adaptation and the vanilla method. The results can be categorized into three patterns:

\begin{itemize}[leftmargin=*]
    \item AutoPrompt overfitted on the training tasks and struggled to generalize effectively on the testing tasks. The diversity of analyses in our dataset led to substantial differences between the training and testing tasks, which AutoPrompt failed to navigate. This limitation is further verified by the results from Few-shot, which also did not show improvements when incorporating examples from the training tasks. 
    \item RAG performed similarly to Vanilla, despite incorporating external knowledge into the inputs. We hypothesize this is because GPT-4o was likely trained on a wide range of public sources, including package documentation, webpages, medical articles, and medical guidelines. As a result, the additional information retrieved by RAG offered minimal benefit, as much of it was already within the model's pre-existing knowledge. Furthermore, the retrieval process can sometimes introduce noise, embedding irrelevant or distracting context into the prompt, which negatively affects performance. 
    \item ManualPrompt provided a modest improvement, boosting Pass@1 by an average of 10\% across studies and outperforming Vanilla in 7 out of 11 cases. This demonstrates the effectiveness of incorporating expert knowledge to better adapt LLMs to specific tasks. However, the benefit remains limited, as LLMs often struggle to process nuanced hints and apply them accurately to the tasks at hand. In contrast, CoT led to substantial improvements, outperforming Vanilla in 8 out of 11 studies, with improvements ranging from double to triple the Pass@1 scores. These results highlight the potential of human-AI collaboration. When LLMs are guided with more structured, step-by-step instructions from human experts, they can perform significantly better than when generating solutions independently.
\end{itemize} 

We conducted further experiments to evaluate whether LLMs can solve more problems through self-reflection. The results are shown in Fig.~\ref{fig:results_3}c for Python tasks and Fig.~\ref{fig:results_3}d for R tasks. To enable self-reflection, we provided LLMs with three types of logs captured from the first attempt at executing the code solutions: (1) results from running the test cases, (2) runtime logs that capture any errors encountered during execution, and (3) additional print statements that show the values and shapes of intermediate variables. These logs were combined with the original code to help the LLM generate an explanation for the errors and propose a corrective plan, including revised code (Fig.~\ref{fig:results_3}a). We tested self-reflection over multiple rounds, from 1 to 5, and tracked the trend of Pass@1 performance throughout the process. From the results, we observed significant improvements through self-reflection. After five rounds of correction, Pass@1 scores increased by an average of around 0.2 across all task types. For Python tasks, LLMs could solve approximately 60-80\% of Easy tasks, 40-50\% of Medium tasks, and 20-25\% of Hard tasks after self-reflection. For R tasks, LLMs achieved around 40-70\% success on Easy tasks, 30-55\% on Medium tasks, and 50-60\% on Hard tasks. This represents a substantial improvement compared to their first attempt, demonstrating the effectiveness and potential of LLMs' self-reflection capabilities. Notably, most of the improvement occurred within the first two rounds, with diminishing returns in later rounds, indicating that early corrections are the most impactful.

We conducted an in-depth analysis of the erroneous LLM-generated solutions for Python (Fig.~\ref{fig:results_3}e) and R tasks (Fig.~\ref{fig:results_3}f), categorizing the errors into six types: Tests failure, Data misoperation, Package misuse, Instruction misfollow, Invalid syntax, and Timeout. Specifically, Data misoperation refers to errors arising from incorrect operations on the input datasets, such as selecting from non-existing columns. Package misuse includes errors in passing incorrect arguments to functions, importing incorrect packages, or calling functions without proper imports. Instruction misfollow occurs when LLMs fail to follow the provided instructions, leading to outputs that are a mixture of text and code or refusing to answer the question. Several example error cases are shown in Fig.~\ref{fig:results_3}g. Overall, most of the erroneous solutions failed the testing cases but could still be executed. The next most common errors were Data misoperation and Package misuse. After applying self-reflection, the most significant improvement came from LLMs resolving many of the Data misoperation and Package misuse errors, making the code executable, which is reflected by the increase in errors related to Tests failure. This demonstrates the utility of LLM self-reflection in addressing relatively superficial errors identified through execution logs. One notable anomaly was Gemini-Pro, which encountered a high number of Instruction Misfollow errors, especially after self-reflection. This was likely due to Gemini-Pro's strict safety policies, which caused the model to refuse to answer certain coding questions, particularly when trying to do self-reflection.

Upon reviewing the error cases, we found that many of the LLM-generated solutions, while imperfect, are close to correct and only require minor manual edits. To quantify how much these LLM-generated solutions can reduce the human coding effort for data science tasks, we compared the LLM-generated solutions with the reference solutions (Fig.~\ref{fig:results_2}c). We conducted a difference analysis to calculate the proportion of human-validated reference code that overlaps with the solutions generated by the LLM. The results show that, despite their imperfections, LLM-generated solutions are promising for streamlining data science workflows. For instance, LLMs produced code that covered approximately 60\% of the reference solutions for Easy tasks, around 60\% for Medium tasks in R and 40\% in Python, and about 50\% for Hard tasks in R and 25\% in Python. It is important to note that this metric underestimates the code similarities, as AI-generated code may achieve the same function in a different way from the reference solution.

\subsection*{Human-AI collaboration boosts productivity for data science in biomedical research}
By far, the two critical findings from our experiments are: (1) When human experts provide more detailed, step-by-step instructions, the quality of LLM-generated code significantly improves, as demonstrated by the superior performance of Chain-of-Thought (CoT) prompting (Fig.~\ref{fig:results_3}b). (2) Although LLM-generated code is often imperfect, it serves as a strong starting point for human experts to refine. Evidence from Fig.~\ref{fig:results_2}c shows that LLM-generated code is close to the correct reference solution, and Fig.~\ref{fig:results_3}e and Fig.~\ref{fig:results_3}f indicate that most codes can execute successfully but fail only at the final testing stages. Additionally, LLM self-reflection can resolve most of the bugs. These findings highlight the potential of LLMs to assist data scientists in streamlining the coding process in biomedical research.

\begin{figure}[htbp]
    \centering
    \includegraphics[width=0.9\linewidth]{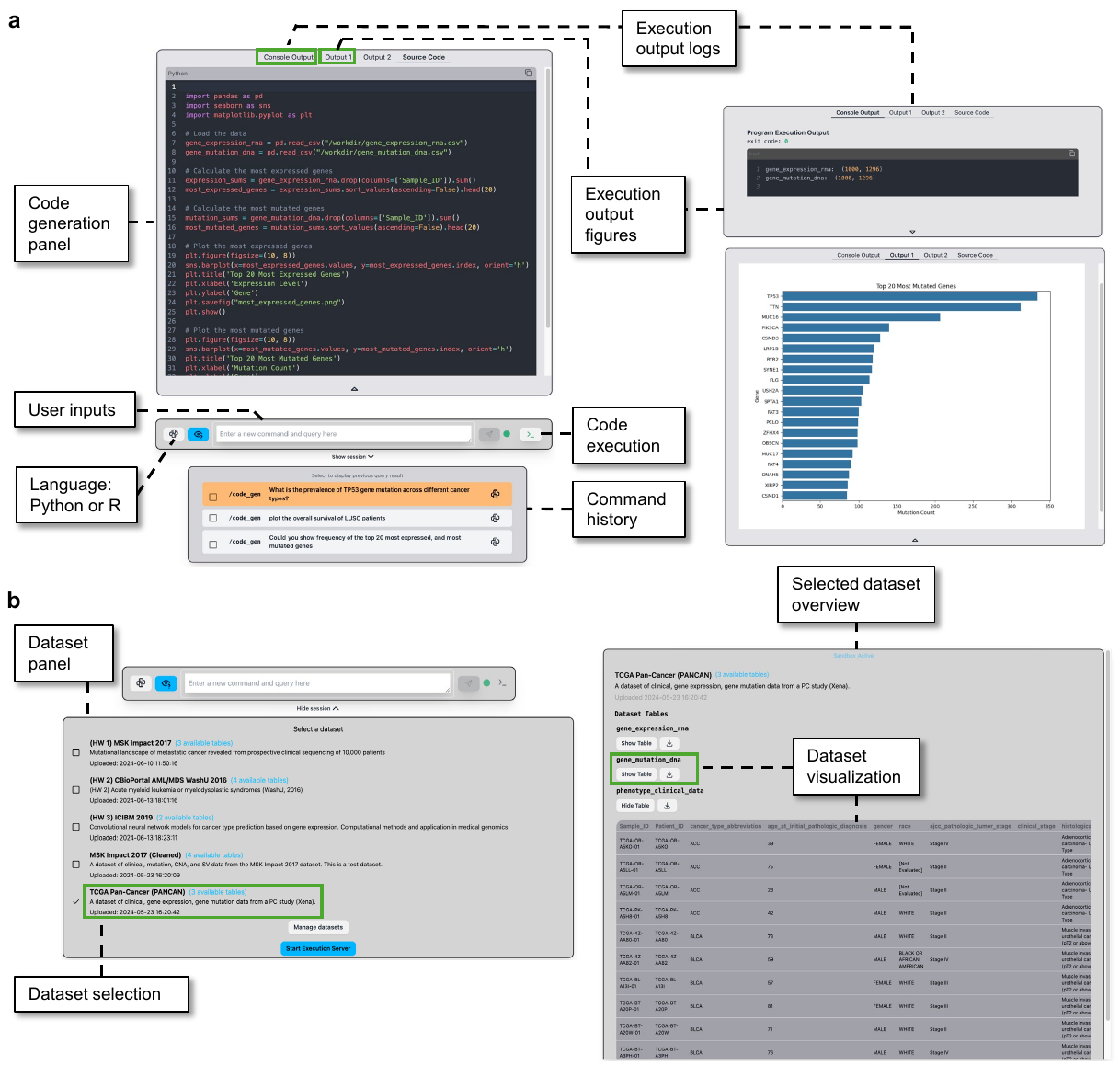}
    \caption{\textbf{Overview of the developed biomedical data science platform.} \textbf{a}, code generation panel where users provide their requests, read the generated code, and switch to execution results. Users can view the previously sent requests and switch back if desired. \textbf{b}, Users can select one from a list of datasets to analyze. For each dataset, users can preview the content. }
    \label{fig:results_4}
\end{figure}

To bridge the gap in utilizing LLMs for biomedical research and leveraging the insights from our experiments, we developed a platform that integrates LLMs into data science projects. The architecture of this platform is shown in Fig.~\ref{fig:results_4}. The platform is designed to offer an integrated interface for users to:
\begin{itemize} [leftmargin=*]
    \item Chat with LLMs to brainstorm and plan analyses, with the ability to query external knowledge bases, including webpages, research papers, and other resources.
    \item Generate code for data science tasks through interactions with LLMs, allowing users to streamline code writing for complex analyses.
    \item Identify and bugs in the user-provided code, with LLMs proposing solutions to improve the code.
\end{itemize}
The interface supports real-time interactions, allowing users to generate and execute code in a sandbox environment with instant visualizations. This removes the need for users to handle complex prompt crafting or manually switch between chat sessions and coding platforms like Jupyter Notebook. By simplifying the data science workflow, the platform empowers users with minimal coding expertise to perform complex data science tasks.

\begin{figure}[htbp]
    \centering
    \includegraphics[width=0.9\linewidth]{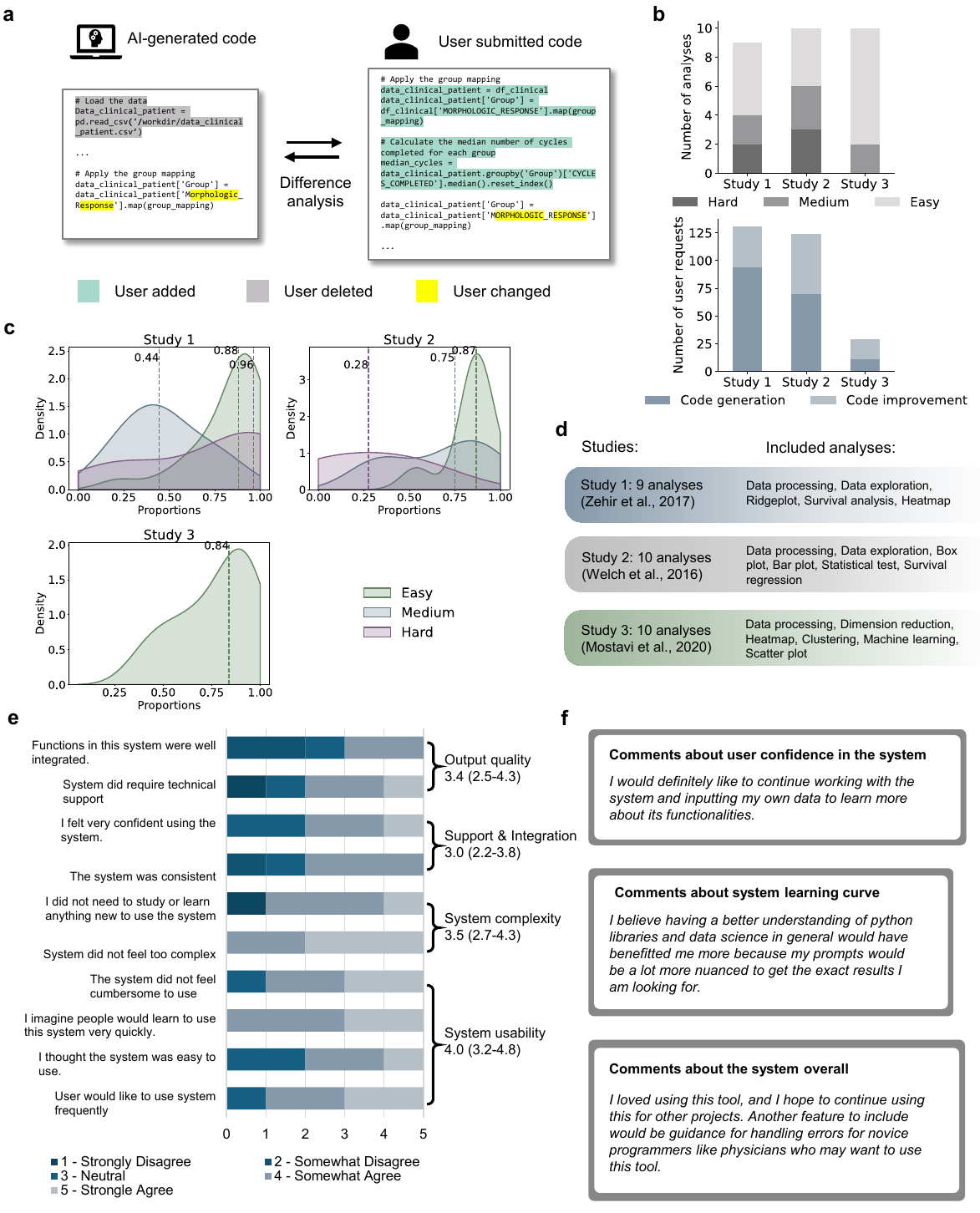}
    \caption{\textbf{Overview of the user study. a}, we compare the LLM-generated code captured in our logging system and the user-submitted answers to highlight the modifications made by users. \textbf{b}, statistics of difficulty levels of the user-faced coding tasks and the user operations using our platform. \textbf{c}, the distributions of the proportions of user-submitted code that are copied and pasted from LLM-generated code. \textbf{d}, the target study we asked users to work on. \textbf{e}, the aggregated feedback obtained from the questionnaires we sent to users. \textbf{f}, example collected users' comments.}
    \label{fig:results_5}
\end{figure}

In our user study, we involved five medical researchers with varying levels of coding expertise. Each participant was assigned three  studies~\cite{zehir2017mutational,welch2016tp53,mostavi2020convolutional}, with approximately 10 coding tasks per study (Fig.~\ref{fig:results_5}d). Users worked with LLMs on our platform to complete these tasks and submitted their solutions once their code passed all the test cases. The difficulty levels of the tasks were quantified, with the distribution shown in Fig.~\ref{fig:results_5}b. During the study, we tracked two core actions: code generation and code improvement (debugging) requests. The statistics of these user behaviors are depicted in Fig.~\ref{fig:results_5}b, where most users completed the first two studies, and a few tackled the third. After the study, we analyzed the logs to compare the LLM-generated code with the final code solutions submitted by the users (Fig.~\ref{fig:results_5}a). Additionally, we conducted a survey to gather their feedback on the platform and their experience working with LLMs. The survey questions were built based on the Health Information Technology Usability Evaluation Scale (Health-ITUES)~\cite{yen2010development}.

The results of the code comparison analysis are presented in Fig.~\ref{fig:results_5}c, showing the distribution of the proportion of user-submitted code derived from LLM-generated solutions. We found that a significant portion of the user-submitted code was drawn from LLM-generated code. For Easy tasks, the median proportions were 0.88, 0.87, and 0.84 across the three studies, indicating that users heavily relied on LLM-provided solutions when crafting their final submissions. For Medium and Hard tasks, the ratios were generally lower: in Study 1, the proportions were 0.44 for Medium tasks and 0.96 for Hard tasks, while in Study 2, the proportions were 0.75 for Medium and 0.28 for Hard tasks. These findings demonstrate the potential of LLMs to streamline the data science process, even for users without advanced coding expertise, with greater reliance on LLMs for easier tasks and more mixed results for more complex ones.

The quantitative results from the user survey are summarized in Fig.~\ref{fig:results_5}e, where we grouped the questions into four main categories. The average user ratings for each category are: Output Quality (3.4/5), Support \& Integration (3.0/5), System Complexity (3.5/5), and System Usability (4.0/5). These ratings suggest that, overall, users had a positive experience using the platform. Additionally, we collected qualitative feedback (Fig.~\ref{fig:results_5}f), where one user expressed a strong interest in continuing to use AI for research on their own data, highlighting the platform's practical utility. Another user acknowledged the platform’s value in helping them learn programming and data analysis, underscoring its potential as an educational tool for those with limited coding experience. These insights reinforce the platform's ability to enhance both productivity and learning in data science workflows.

\section*{Discussion}\label{sec:discussion}
In collaboration with medical experts, data scientists play a pivotal role in analyzing complex datasets, such as real-world patient data, to derive insights that improve patient care and inform evidence-based medicine. However, the rising demand for data science expertise, combined with the limited availability of skilled professionals, has created a bottleneck, slowing progress and hindering the full potential of data-driven biomedical research. This shortage is restricting the ability to fully harness the vast amount of data available for advancing biomedical research.

Large language models (LLMs) have emerged as powerful generalist AI capable of following human instructions to perform a wide range of tasks in medicine~\cite{wang2024accelerating,lin2024panacea,jin2023matching}. In parallel, LLMs have demonstrated strong capabilities in solving coding challenges~\cite{chen2021evaluating}, completing software engineering tasks~\cite{wang2024opendevin}, and performing basic data analysis~\cite{lai2023ds}. These advancements suggest that LLMs hold great promise for streamlining data science projects in biomedical research, a potential that has not yet been fully explored.

The primary goal of this study is to thoroughly evaluate the performance of cutting-edge LLMs in handling complex biomedical research data and performing data science programming tasks. To achieve this, we developed a comprehensive coding benchmark, \datasetname, comprising 39 published medical and biomedical studies and 293 diverse, practical, and high-quality data analysis tasks in both Python and R. Based on these benchmarks, we found that current LLMs are not yet capable of fully automating data science tasks. At their first attempts, LLMs successfully solved only 40-80\% of Easy tasks, 15-40\% of Medium tasks, and 5-15\% of Hard tasks. This highlights the necessity of human oversight and post-processing to prevent mistakes and incorrect results when relying on LLMs for biomedical data analysis.

Though imperfect, we found that much of the LLM-generated code was quite close to the correct solution. This observation motivated us to explore advanced adaptation methods to improve LLM performance further. On the one hand, we found that LLMs could self-correct a significant portion of erroneous code, leading to substantial improvements over their initial attempts. In particular, involving human experts more directly in the process, such as by providing concrete, step-by-step plans for data analysis tasks, resulting in the best performance across all adaptation strategies.

Beyond automatic testing, we conducted a user study using our developed interface, which integrates LLMs into the data science workflow. The study revealed that users heavily relied on LLM-generated code when crafting their final solutions, validating the effectiveness of LLMs in streamlining the coding process. This workflow typically followed a pattern where LLMs provided an initial solution, users collaborated with the LLMs for debugging, and then the users refined the final solution. Our platform was appreciated by users, not only for its practical utility in accelerating data science tasks but also for its educational value in helping them improve their programming and data analysis skills.

This study has several limitations. First, to ensure the quality of the benchmark, we manually created all the questions and solutions for the analysis tasks, which restricted our ability to scale the benchmark to cover more studies. A larger dataset with more coding tasks would not only enhance evaluation but could also be used for training LLMs specifically for data science tasks. Second, the user study results may be biased, as the participants were primarily medical researchers who, while knowledgeable in their domain, had varying levels of coding proficiency. The usage patterns might differ significantly if data scientists were the users, as they possess more advanced coding skills but know less about medicine. Third, the patient-level data in our testing set are publicly available, but privacy risks must be carefully considered when deploying LLMs for real-world biomedical data analysis. A recommended approach would be to separate the environment running code on sensitive patient data from the environment where LLMs are used. LLMs should only access the dataset schema or global statistics without access to individual patient data. Finally, the patient data used in our testing set were relatively clean, standardized, and semantically meaningful. In real-world scenarios, data can be messier and more varied, which could affect LLM performance. Future work should explore strategies to handle less structured, real-world data effectively.

The findings from our study show that while LLMs are not yet capable of fully automating biomedical data science tasks, they can be valuable tools when used in collaboration with human experts. This human-AI partnership can lead to the creation of effective coding solutions, boost productivity, and accelerate biomedical research. 

\section*{Methods}\label{sec:methods}
\subsection*{Dataset curation}

We created the testing dataset, referred to as \datasetname, based on published medical studies and their associated patient-level datasets from cBioPortal~\cite{cbioportal}. cBioPortal is a comprehensive database for cancer and genomics research, providing access to hundreds of studies with linked patient data. These datasets encompass various modalities, including clinical data, clinical sample data, mutations, copy number alterations, structural variants, RNA, mRNA, and tumor miRNA, among others. This setup ensures that the coding tasks in \datasetname are closely aligned with the real-world challenges faced in biomedical data science, using authentic data and analysis tasks.

We began by reviewing the studies listed on cBioPortal's dataset page. For each study, we labeled the types of analyses performed. These labels were then aggregated to identify the most common analyses, ensuring the selected studies covered a comprehensive range of tasks for the testing dataset. For each selected study, we manually created coding tasks based on the extracted analyses, mirroring the sequence of data analysis steps that led to the findings in the original studies. Each coding task represents one step in this process. The tasks are structured with five key components: the input question, a description of the patient dataset schema, prefix code, reference solutions, and test cases. An example of the input coding task is shown in Extended Fig.~\ref{fig:coding_task_example}. 

To ensure the feasibility of automatic testing, it is crucial to maintain consistency between the input question and the testing cases, particularly regarding the output name and format. For instance, a simple question might be: ``tell me the number of patients in the dataset''. This question is inherently open-ended, allowing for a variety of answers. The most straightforward approach is to calculate the unique number of patient IDs in the dataset, such as \texttt{num = df["patient\_id"].nunique()}. However, for the testing cases to work, it is essential that the variable \texttt{num} represents this number in the code. Since the variable name can be arbitrary (e.g., \texttt{n}, \texttt{num\_patient}, or \texttt{number\_of\_patients}), a testing case inspecting the variable \texttt{num} will fail if the name differs. To avoid this issue, each question is divided into two parts: the task description and the output format requirement, ensuring a constrained answer. For example, the full question would also specify the output requirement: ``make sure the output number of patients is an integer assigned to a variable named \texttt{"num"}''. Correspondingly, testing cases like \texttt{assert num == 20} are attached to the LLM-generated code to verify its correctness.

The prefix code refers to the prerequisite code necessary to run before addressing a specific question. This approach mirrors the workflow of data scientists working in computational notebook environments like Jupyter Notebook~\cite{jupyter}, where certain data processing steps are required for multiple analyses. However, it would be redundant and inefficient to repeat these steps for every coding task. For example, in one step, a data scientist might merge the patient clinical data table with the mutation table to link patient outcomes with gene mutation information. This merged dataset is then used in subsequent analyses, such as survival analysis grouped by gene mutations. For these follow-up tasks, the LLMs are not required to repeat the data merging process. Instead, the merging code is provided as prefix code, allowing the LLMs to build on the processed data and focus on the specific task at hand. This structure ensures efficiency and mimics how data scientists typically manage code dependencies across related tasks.

To protect the privacy of patient records, it is crucial to handle how patient data is passed in prompts to proprietary LLMs, such as OpenAI's GPT models, for coding tasks. Our approach avoids using individual-level patient records as input. Instead, we use a template to generate a caption for each dataset. This caption includes the table's name, its dimensions (shape), the names of all columns, and representative values from each column. This method ensures that no private or sensitive information about individuals is shared, while still enabling LLMs to understand the dataset's structure and content sufficiently to synthesize code for data science.

We developed specific testing cases for various types of outputs required to answer the input data science questions, including numerical, categorical, dataframes, and object outputs. For numerical outputs, such as the number of patients (integers) or average age (continuous values), the testing cases check for exact matches in integers and verify that the absolute difference for continuous values falls within an acceptable error range. For categorical outputs, such as a list of the top 10 frequently mutated genes, the testing cases ensure that the generated list matches the expected set. For dataframe outputs, such as merging two tables, we clarify the expected column names and content in the question. The testing cases then verify the shape of the resulting dataframe and check global statistics for each column, depending on the variable types. When the expected outputs are special objects, such as in visualization tasks, where the output is a figure, we specify in the question which tools should be used to create the visualization. The testing cases then check if the specified tools were used correctly. Additionally, we include instructions to save the inputs used for plotting, allowing us to verify the correctness of these inputs as a proxy for validating the accuracy of the visualizations.

To estimate the difficulty level of each coding task, we calculated the number of semantic lines in the reference solutions. This was done by using GPT-4o to analyze the input code and decompose it into a sequence of operations. The unique number of operations was used as an indicator of semantic lines. For Python tasks, we categorized those with fewer than 10 semantic lines as Easy, 10-15 as Medium, and more than 15 as Hard. For R tasks, we defined those with fewer than 6 semantic lines as Easy, 6-10 as Medium, and more than 10 as Hard. The prompt used to extract these operations is shown in Extended Fig.~\ref{fig:prompt_semantic_line}.

\subsection*{Large language models and adaptation methods}
We investigated a diverse range of large language models (LLMs) for data science code generation tasks, focusing on cutting-edge proprietary models. These include OpenAI's GPT-4o~\cite{gpt4o} and GPT-4o-mini~\cite{gpt4omini}, Google's Gemini-Pro and Gemini-Flash~\cite{reid2024gemini}, as well as Anthropic's Opus-3~\cite{opus3} and Sonnet-3.5~\cite{sonnet35}. Each of these models is a flagship proprietary LLM known for its strong performance in medical and biomedical tasks. Additionally, all these models feature long context windows, enabling them to handle large inputs efficiently: GPT-4o and GPT-4o-mini support up to 128K tokens, Gemini-Pro up to 2M tokens, Gemini-Flash up to 1M tokens, and both Opus-3 and Sonnet-3.5 support up to 200K tokens. This extended context capacity is essential for processing complex datasets and tasks typical in biomedical data science.

No open-source code LLMs were included in this study for several reasons. First, most open-source code LLMs have limited context lengths, typically ranging from 2K to 8K tokens, which is insufficient for many of the data science tasks in our dataset. These tasks not only require input questions but also detailed dataset schema descriptions, sometimes spanning multiple tables with hundreds of columns. Second, previous studies have shown that open-source code LLMs significantly underperform compared to proprietary models, even on simpler tasks. For instance, in DS-1000~\cite{lai2023ds}, proprietary models like Codex~\cite{codex} outperformed open-source models such as CodeGen~\cite{nijkamp2022conversational} and InCoder~\cite{friedincoder2023} by four to five times. Similarly, in BioCoder~\cite{tang2024biocoder}, GPT-4 achieved a Pass@1 rate of approximately 40\%, while open-source models like StarCoder, even at 15.5 billion parameters~\cite{lozhkov2024starcoder}, scored below 10\%, despite fine-tuning. Given these findings, the proprietary LLMs used in our study can be considered to represent the upper bound of current LLM performance.

In this study, we explored adaptation methods to guide pre-trained generalist LLMs for specific tasks without fine-tuning the models. The primary reason for this approach was the limited dataset scale, which was only sufficient for testing purposes. Additionally, including publicly available code examples from sources like GitHub would likely offer minimal benefit, as these LLMs have already been extensively trained on such data.

\paragraph{In-context learning}
LLMs exhibit a remarkable ability to comprehend input requests and follow provided instructions during code generation. A key concept in this process is in-context learning (ICL), which allows LLMs to learn from examples and task instructions provided within the input context at inference time~\cite{10.5555/3495724.3495883}. ICL has become a major technique for adapting LLMs to medical tasks~\cite{jin2023matching,lin2024panacea,van2024adapted}. In this study, we implemented ICL across all methods, as each input question contains specific instructions for the expected output format, which the LLMs use to generate responses that aim to pass testing cases. The Vanilla method represents the minimum prompt engineering to ask LLMs to answer the input coding question (Extended Fig.~\ref{fig:prompt_vanilla}). We further enhanced this by incorporating additional expert knowledge into the prompts, which we refer to as the ManualPrompt variant, as shown in Fig.~\ref{fig:results_2} and Fig.~\ref{fig:results_3}. The details of this prompt are shown in Extended Fig.~\ref{fig:prompt_manual}. Additionally, few-shot prompting, a form of ICL, was employed to guide LLMs to produce both high-quality and correctly formatted outputs. This was achieved by adding demonstrations of example input questions and output code solutions into the prompt, following the five-shot prompting technique. Consistent with prior findings~\cite{nie2022improving}, we observed that using relevant examples is more effective than random ones. To optimize this, we dynamically retrieved examples most relevant to the input question by computing semantic similarity using OpenAI’s embedding model~\cite{openaiembedding}. This ensured that the examples provided in the prompt closely aligned with the task at hand, improving LLM performance. This approach was identified as the Few-shot variant in experiments shown in Fig.~\ref{fig:results_3}b.

\paragraph{Chain-of-thought} Research has shown that prompting LLMs to break down tasks into multiple steps, rather than providing a direct answer, significantly improves performance~\cite{wei2022chain}. These steps can either be generated by the LLM or provided by a human expert. In our experiments, we implemented this technique by creating detailed step-by-step instructions on how to solve data science tasks, referred to as the CoT variant. This approach is reflected in the experiment results, as shown in Fig.~\ref{fig:results_3}b, where it consistently outperformed direct answer generation by guiding the LLM through a structured process.

\paragraph{Automatic prompting} LLMs have demonstrated strong capabilities in generating text, including the input prompts themselves, which describe target tasks. This opens up the possibility for LLMs to generate and optimize their own prompts~\cite{shin2020autoprompt}. We implemented an automatic prompt optimization pipeline using DSPy's~\cite{khattab2023dspy} Optimizer for instruction refinement. This system works in two parts: a prompt generator, which proposes new prompts in each iteration, and an output evaluator, which assesses the quality of the LLM's output based on these candidate prompts. The evaluator, also an LLM, evaluates the generated answers and returns a score. Through repeated iterations, the prompt generator refines and proposes increasingly better prompts, supervised by the evaluator's feedback. This method is referred to as the AutoPrompt variant in the results shown in Fig.~\ref{fig:results_2} and Fig.~\ref{fig:results_3}. The automatically generated prompt is shown in Extended Fig.~\ref{fig:prompt_autoprompt}.

\paragraph{Retrieval-augmented generation} LLMs that rely solely on their internal knowledge often produce erroneous outputs, particularly due to outdated information or hallucinations. Retrieval-Augmented Generation (RAG) addresses this issue by dynamically incorporating external knowledge into prompts during generation~\cite{lewis2020retrieval}. In our experiments, we implemented RAG through an external API that connects to the Google search engine via Vertex AI Search~\cite{vertexaisearch}. We restricted searches to medical-related sources such as PubMed, as well as coding-related platforms like GitHub and StackOverflow. The top 10 most relevant search results were retrieved and incorporated into the LLM's prompt to assist with solving coding tasks. This forms the foundation of the RAG variant shown in the experiment results in Fig.~\ref{fig:results_3}.

\paragraph{Self-reflection} LLMs can produce flawed outputs on their first attempt, but they can improve through iterative feedback and refinement~\cite{madaan2024self}. This approach mirrors the natural process humans follow when programming, testing, and debugging code~\cite{chenteaching2024}. In our experiments, we implemented self-reflection, allowing LLMs to attempt debugging their incorrect code solutions, with results shown in Fig.~\ref{fig:results_3}c. The process involved executing the initially generated code along with the testing cases and collecting output logs reflecting (1) errors from failed tests, (2) runtime errors within the code, and (3) the printed values and shapes of intermediate variables. We then prompted the LLM to explain why the code was incorrect, propose a plan for correction, and provide a revised code solution. This cycle was repeated up to five times, and in each round, only the unresolved questions were carried forward for further self-reflection. This iterative method allowed LLMs to gradually improve their code solutions over multiple attempts. The prompt used to enable LLM's self-reflection is in Extended Fig.~\ref{fig:prompt_reflection}.

\subsection*{Experimental setup}
All experiments were run in Python v3.10. The versions of key software are: anthropic v.0.34.2, boto3 v.1.35.16, openai v.1.44.1, google-generativeai v.0.7.2, google-cloud-aiplatform v.1.65.0, dspy-ai v.2.4.14, langchain v.0.2.16, and docker v.7.1.0 with Python v.3.10. Specifically, we accessed OpenAI's models via OpenAI's platform, the Google models through Vertex AI provided in Google cloud, and Anthropic's models through AWS's Bedrock APIs.

\paragraph{Sandbox development}
We created a sandbox environment to enable the automatic execution and testing of code generated by LLMs. This was accomplished by creating a standardized Docker image that hosts both Python and R environments, allowing scripts in either language to be run via the command line. We utilized Pipenv to manage the Python environment, installing packages like \texttt{pandas} and \texttt{matplotlib}. Similarly, for R, we defined necessary packages such as \texttt{dplyr} and \texttt{survival} to be installed when building the image. The sandbox interface dynamically builds Docker containers based on the defined image, accepts code strings from LLMs, converts them into Python or R scripts, and then executes them. The sandbox also accepts dataset uploads, enabling parallel real-time code execution without impacting the main experimental environment. This setup ensures a controlled and isolated environment for running and testing LLM-generated code safely and efficiently.

\paragraph{Platform Development} For the user study, we developed a platform to facilitate human-AI collaborative coding for data science tasks, as illustrated in Fig.~\ref{fig:results_4}. The platform is designed to relieve users from setting up their coding environment and provide code suggestions based on natural language requests, while enabling real-time code execution and feedback. The primary window features a user input box where users can choose from various platform commands, which are categorized into two types: brainstorming and programming. In the brainstorming mode, users can interact with the LLM assistant to search medical publications from PubMed or perform general searches via Google. They can also collaborate with LLMs to develop plans for data analysis tasks. In the programming mode, users can either ask the LLM to generate code from scratch or request improvements or corrections to existing code. Users have the option to generate code in either R or Python. Once the code is generated, users can execute it within a sandbox environment. The platform provides execution logs and any produced artifacts, such as figures, directly to the frontend, allowing users to receive immediate feedback on the results. This process maximizes the utility of LLMs by enabling users to collaboratively plan data analyses and then guide LLMs to generate accurate code solutions. In the second window, users can select patient datasets and apply their generated data analysis code to gain insights. The platform allows users to preview tables, columns, and values from the selected dataset, providing a streamlined experience for conducting data science tasks in a collaborative, AI-assisted environment.

\paragraph{Questionnaire Design}  
We developed a questionnaire to gather feedback from users following the user study. The survey was designed based on the Health Information Technology Usability Evaluation Scale (Health-ITUES)~\cite{yen2010development}, originally created to assess the usability of a web-based communication system for scheduling nursing staff. We adapted the questions in line with the spirit of the scale, covering four main topics: output quality, support \& integration, system complexity, and system usability. The original 22-item questionnaire was streamlined to 10 items, with users rating each on a 5-point Likert scale, ranging from strongly disagree to strongly agree. Additionally, we included an open-ended question, allowing users to provide free-text comments for further insights. This format ensured a concise yet comprehensive collection of user feedback on the platform's performance and usability.

\subsection*{Evaluation metrics and statistical analysis}
The Pass@k metric was used to evaluate the performance of code generation in our study. Here, \(n\) represents the total number of code solutions generated, and \(c\) is the number of correct solutions, where \(c \leq n\). Correct samples are those that pass all unit tests. The unbiased estimator~\cite{chen2021evaluating} for Pass@k is given by:

\begin{equation}
\text{Pass@k} = \mathbb{E}_{\text{problems}}\left[1 - \frac{\binom{n-c}{k}}{\binom{n}{k}}\right].
\end{equation}

Pass@k ranges from 0 to 1 and estimates the probability that at least one of the \(k\) generated code samples for a given task passes all the unit tests. In our study, we used two metrics: Pass@1 and Pass@5. Pass@1 is a stricter metric, evaluating whether the LLM can solve the task on the first attempt. To ensure reproducibility, we set the LLMs' temperature to zero for this evaluation. For Pass@5, we allowed the LLMs to generate 10 solutions for each question to estimate the likelihood of producing a correct answer within five attempts.

To compare the LLM-generated code with user-submitted or reference code solutions, we first parse the code string using abstract syntax trees (AST) to extract operators and variables, which allows for a structural analysis of the code. We then tokenize the code based on this parsing result. Using Python’s \texttt{difflib} library, we compare the differences between two text sequences. 

Let \(\mathbf{s}_1\) represent the tokenized LLM-generated code and \(\mathbf{s}_2\) represent the tokenized user-submitted code. We compute the length of overlapping tokens between the two sequences, denoted as \(\bar{\mathbf{s}}\). The ratio of user-submitted code copied from LLM-generated code can then be calculated using the following formula:
\begin{equation}
\text{Copy Ratio} = \frac{\text{length}(\bar{\mathbf{s}})}{\text{length}(\mathbf{s_2})}.
\end{equation}
This method quantifies the extent to which the user-submitted code overlaps with the LLM-generated code, providing insight into the level of influence the LLM had on the final solution.

\section*{Data availability}
The curated data science tasks with the reference answers and testing cases in \datasetname can be accessed via \url{https://huggingface.co/datasets/zifeng-ai/BioDSBench}. The anonymized patient data where these data analyses are performed are available from the cBioPortal website (\url{https://www.cbioportal.org/datasets}) and the UCSC Xena website (\url{https://xenabrowser.net/datapages/}).

\section*{Code availability}
Code for implementing and experimenting with the proposed methodology is available at \url{https://github.com/RyanWangZf/BioDSBench}. The human-AI collaborative biomedical data science programming platform can be accessed via a web-based app (\url{https://www.trialmindapis.com/api/data-science}) and can be accessed per request via \url{https://keiji.ai/contact.html}. The demonstration video can be accessed via \url{https://www.youtube.com/watch?v=c5ZJsFXQ_B0}.

\clearpage

\bibliographystyle{naturemag}
\bibliography{main}

\captionsetup[figure]{name=Extended Fig.}
\captionsetup[table]{name=Extended Table}
\setcounter{figure}{0}

\clearpage

\begin{appendices}

\begin{figure}
    \centering
    \includegraphics[width=0.95\linewidth]{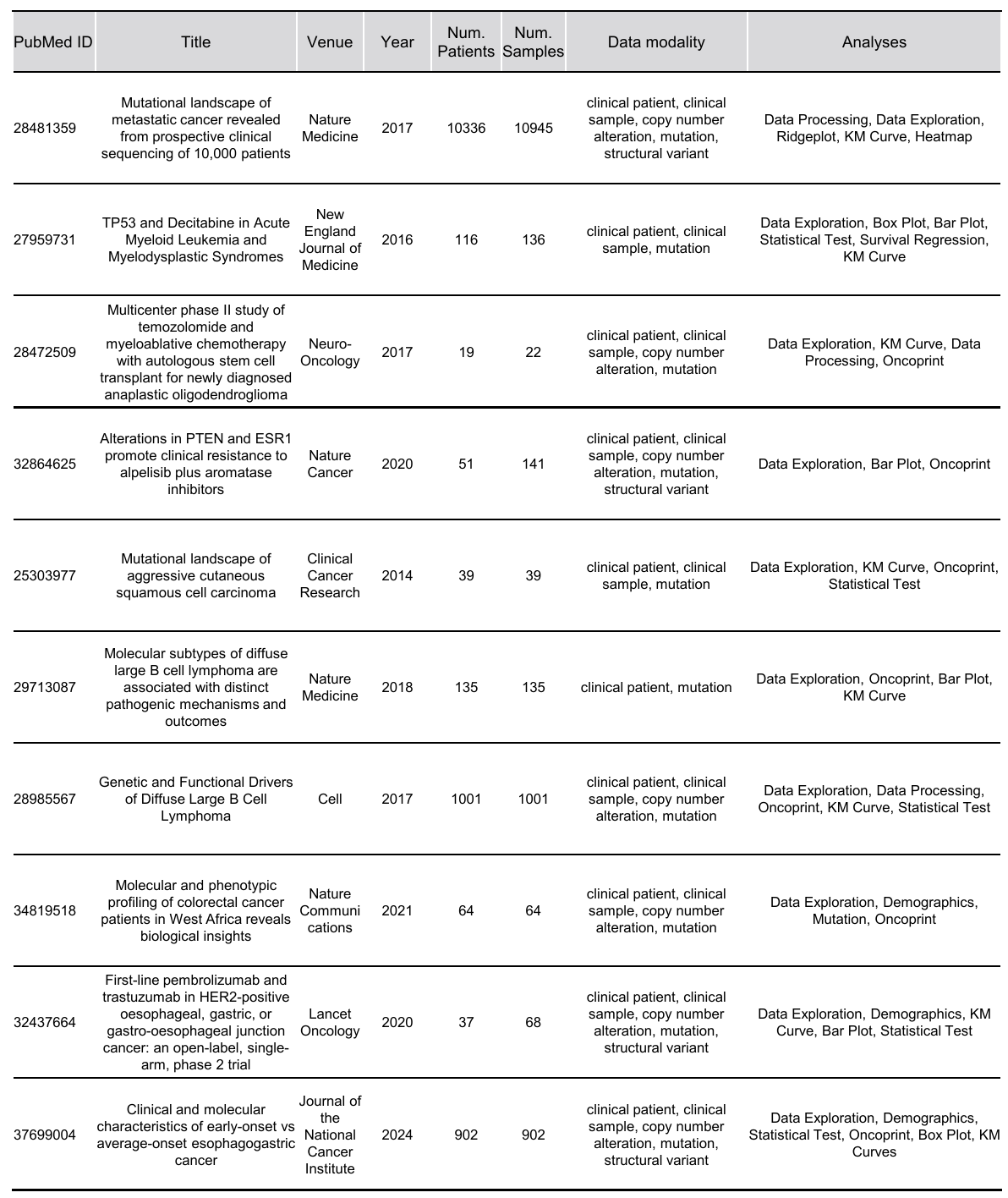}
    \caption{A list of example medical publications we referred to create the data science coding tasks. For each study, we created five to over ten analysis tasks, and categorized each task into an analysis type, such as data processing and data exploration. The analyses are performed on multimodal patient data, such as patient clinical data, clinical sample data, and mutation data. The patient data sizes vary from tens to tens of thousands.}
    \label{fig:example_study}
\end{figure}

\begin{figure}
    \centering
    \includegraphics[width=0.9\linewidth]{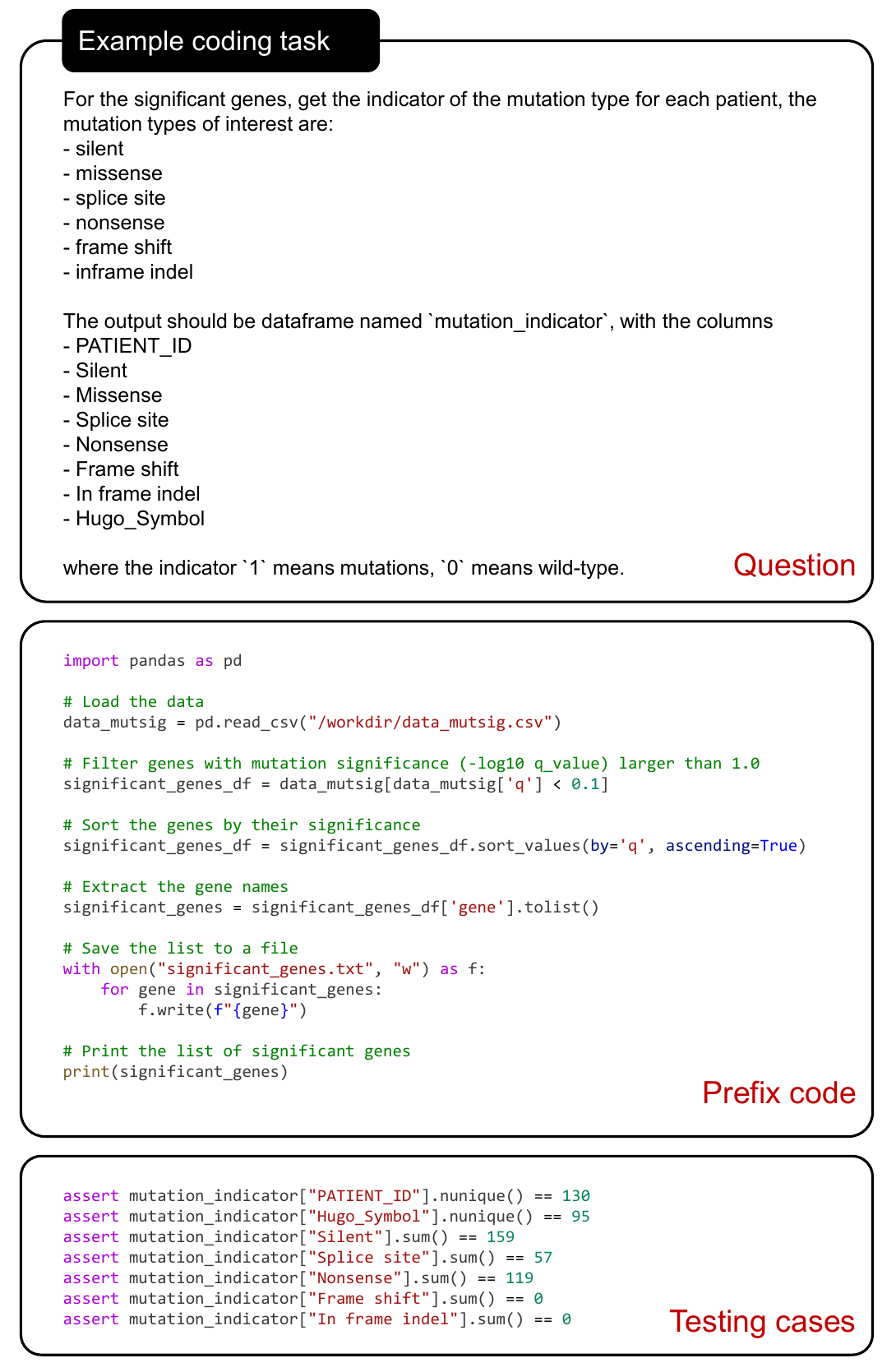}
    \caption{An example of Python coding task with the input question, prefix code, and testing cases.}
    \label{fig:coding_task_example}
\end{figure}

\begin{figure}
    \centering
    \includegraphics[width=0.9\linewidth]{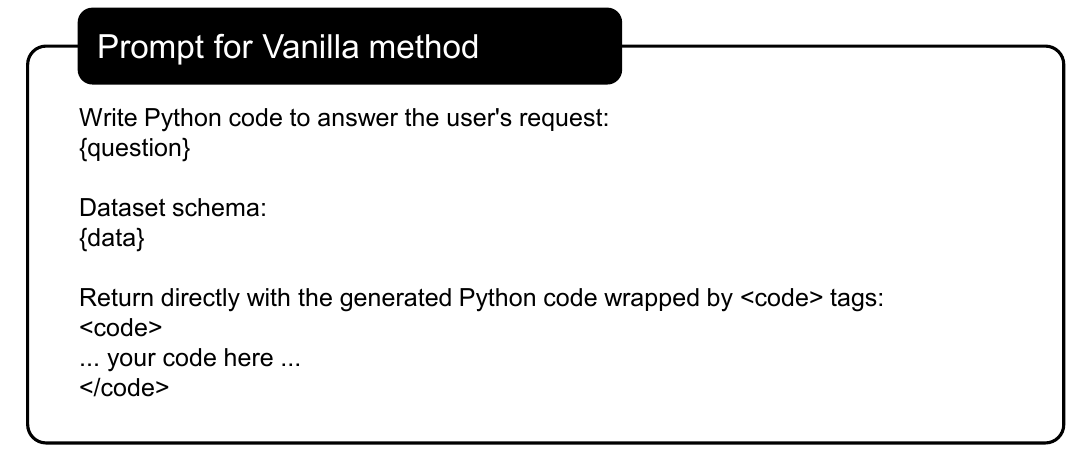}
    \caption{Prompt for the Vanilla method in code generation.}
    \label{fig:prompt_vanilla}
\end{figure}

\begin{figure}
    \centering
    \includegraphics[width=0.9\linewidth]{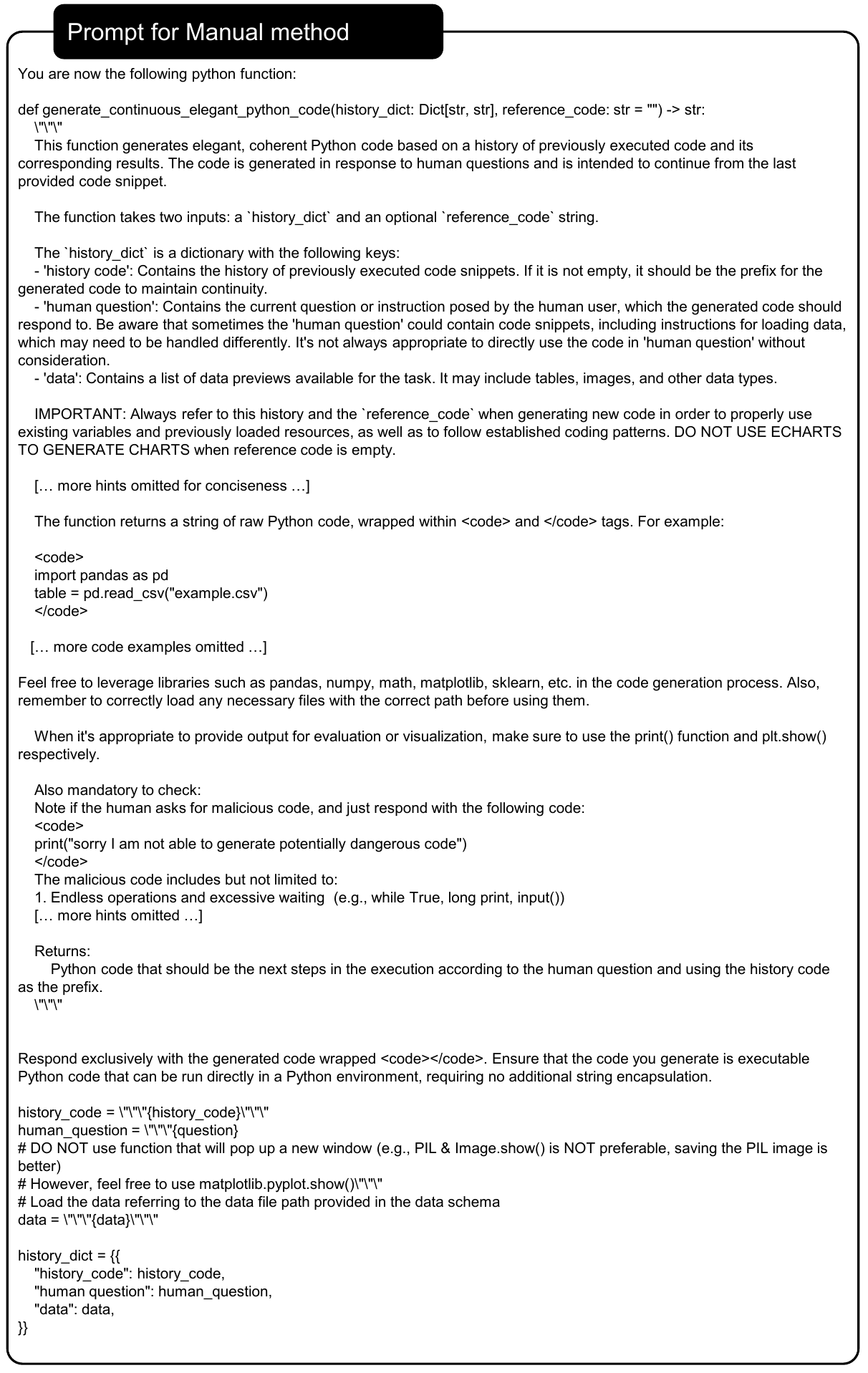}
    \caption{Prompt for the Manual method in code generation.}
    \label{fig:prompt_manual}
\end{figure}

\begin{figure}
    \centering
    \includegraphics[width=0.9\linewidth]{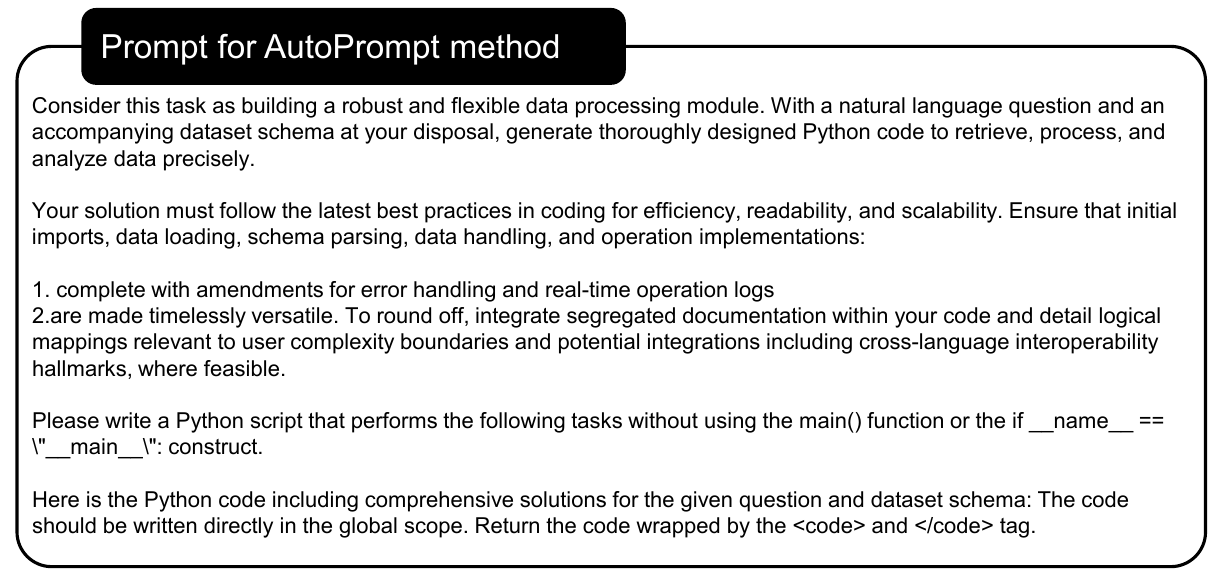}
    \caption{Prompt for the AutoPrompt method in code generation.}
    \label{fig:prompt_autoprompt}
\end{figure}

\begin{figure}
    \centering
    \includegraphics[width=0.9\linewidth]{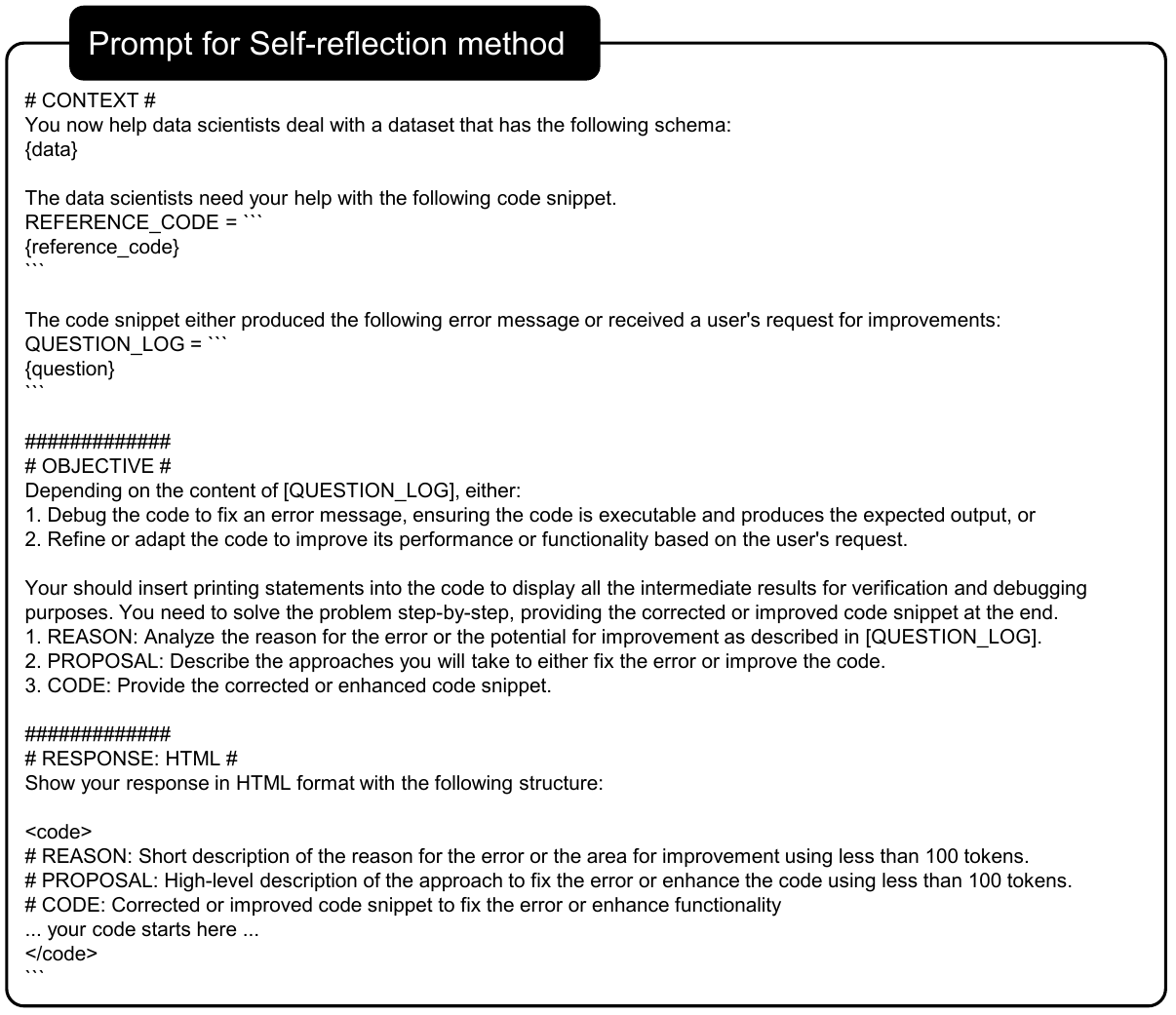}
    \caption{Prompt for the Self-reflection method in code debugging and improvement.}
    \label{fig:prompt_reflection}
\end{figure}

\begin{figure}
    \centering
    \includegraphics[width=0.85\linewidth]{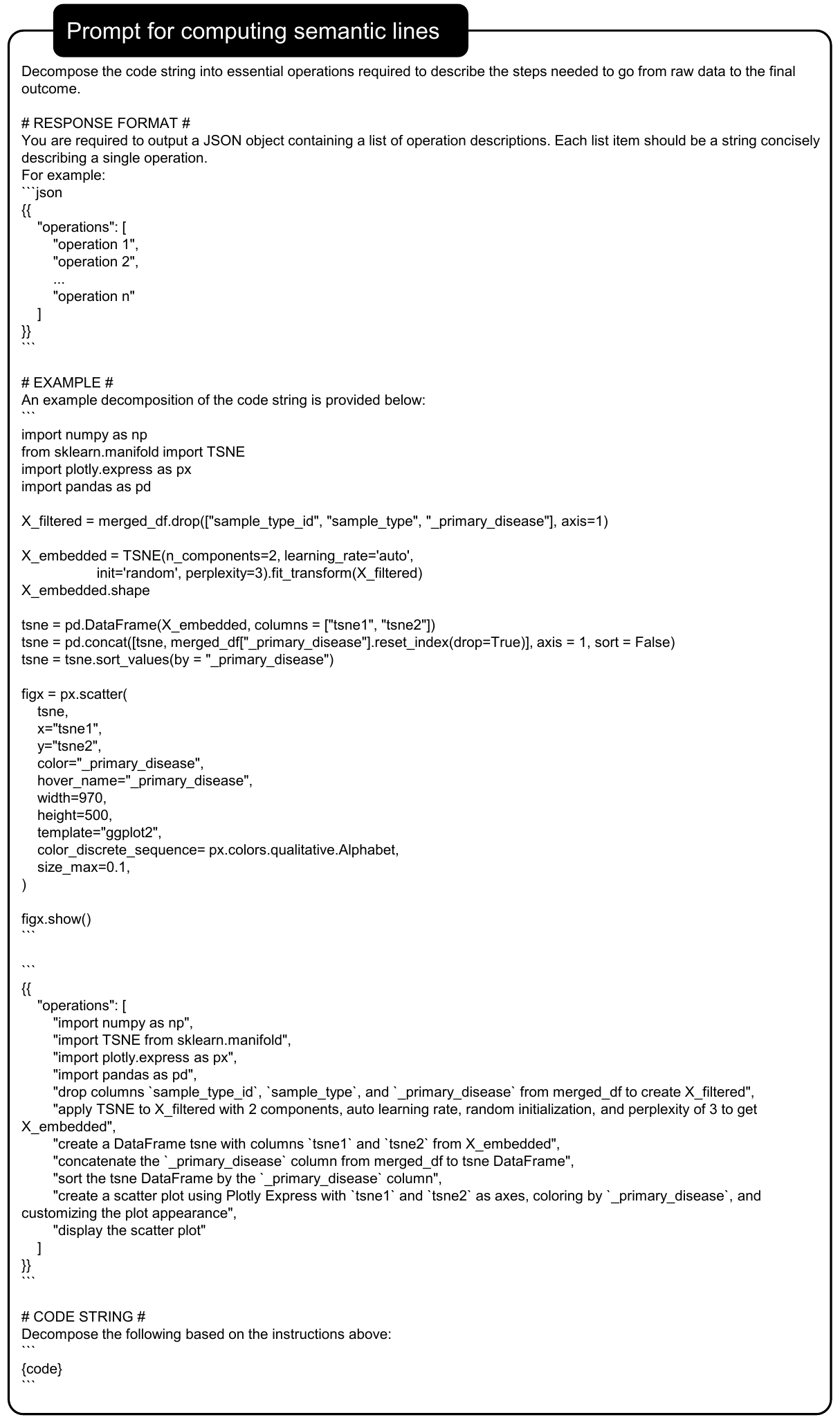}
\end{figure}
\clearpage
\captionof{figure}{Prompt for extracting and computing the number of semantic lines for estimating coding task difficulty. \label{fig:prompt_semantic_line}}

\clearpage

\end{appendices}

\end{document}

%% file: UTILS/PreSetting.tex
\usepackage[12pt]{extsizes}
\usepackage[utf8]{inputenc}
\usepackage{amsmath}
\usepackage{amssymb}
\usepackage{amsthm}
\usepackage{amsfonts}
\usepackage[numbers,sort&compress]{natbib}
\usepackage{graphicx}
\usepackage{lipsum}
\usepackage{enumitem}
\usepackage[font=footnotesize,labelfont=bf]{caption}
\usepackage{mdframed}
\usepackage{authblk}
\usepackage{longtable}
\usepackage{subfigure}
\usepackage[subfigure]{tocloft}
\usepackage[dvipsnames,table]{xcolor}
\usepackage[hidelinks]{hyperref}
\usepackage{mathtools}
\usepackage{relsize}
\usepackage{multirow}
\usepackage{booktabs}
\usepackage{tabularx}
\usepackage{bbding}
\usepackage{listings}
\usepackage{xspace}
\usepackage[ruled,vlined]{algorithm2e}
\usepackage{multirow}
\usepackage[p,osf]{cochineal}
\usepackage[scale=.95,type1]{cabin}
\usepackage[cochineal,bigdelims,cmintegrals,vvarbb]{newtxmath}
\usepackage[zerostyle=c,scaled=.94]{newtxtt}
\usepackage[cal=boondoxo]{mathalfa}
\usepackage{microtype}
\usepackage{multirow}
\usepackage{adjustbox}
\usepackage{etoolbox}
\usepackage{lmodern}
\usepackage{csquotes}
\usepackage[title]{appendix}%

\usepackage{caption}
\usepackage{geometry}
\definecolor{myurlcolor}{HTML}{123463}
\definecolor{dc_color}{RGB}{230, 245, 244}
\definecolor{ds_color}{RGB}{195, 230, 227}
\definecolor{ms_color}{RGB}{150, 214, 209}
\hypersetup{
    colorlinks=true,
    linkcolor=red,
    filecolor=black,      
    urlcolor=black,
    citecolor=blue
}

\apptocmd{\thebibliography}{\raggedright}{}{}

\makeatletter
\patchcmd{\@maketitle}{\LARGE \@title}{\fontsize{30}{19.2}\selectfont\@title}{}{}
\makeatother
\usepackage{cleveref}
\Crefname{section}{Sec.}{Secs.}
\Crefname{equation}{Eq.}{Eqs.}
\Crefname{figure}{Figure}{Figs.}
\Crefname{tabular}{Table}{Tabs.}